%% file: main.tex
\documentclass{sigkddExp}
\pdfoutput=1

\input{packages.tex}

\graphicspath{ {./images/} }

\def\cw#1{\textcolor{black}{#1}}

\begin{document}
%

\title{Classification of multivariate weakly-labelled time-series with attention}
%

\numberofauthors{2}
%


\author{
%
\alignauthor Surayez Rahman \\
       \affaddr{Faculty of Information Technology}\\
       \affaddr{Monash University, Clayton Campus}\\
       \affaddr{Melbourne, Australia}\\
       \email{srah0001@student.monash.edu}
\alignauthor Chang Wei Tan\\
       \affaddr{Faculty of Information Technology}\\
       \affaddr{Monash University, Clayton Campus}\\
       \affaddr{Melbourne, Australia}\\
       \email{chang.tan@monash.edu}
}

\date{06 November 2020}

\maketitle
\begin{abstract}
\cw{This research identifies a gap in weakly-labelled multivariate time-series classification (TSC), where state-of-the-art TSC models do not perform well.
Weakly labelled time-series are time-series containing noise and significant redundancies.}
In response to this gap, this paper proposes an approach of exploiting context relevance of subsequences from previous subsequences to improve classification accuracy. 
To achieve this, state-of-the-art Attention algorithms are experimented in combination with the top CNN models for TSC (FCN and ResNet), in an CNN-LSTM architecture. 
Attention is a popular strategy for context extraction with exceptional performance in modern sequence-to-sequence tasks.
This paper shows how attention algorithms can be used for improved \cw{weakly labelled} TSC \cw{by evaluating models on a multivariate EEG time-series dataset obtained using a commercial Emotiv headsets from participants performing various activities while driving.}
These time-series are segmented into subsequences and labelled to allow supervised TSC.

\end{abstract}


\section{Introduction}
\begin{figure}[t]
\includegraphics[width=\columnwidth]{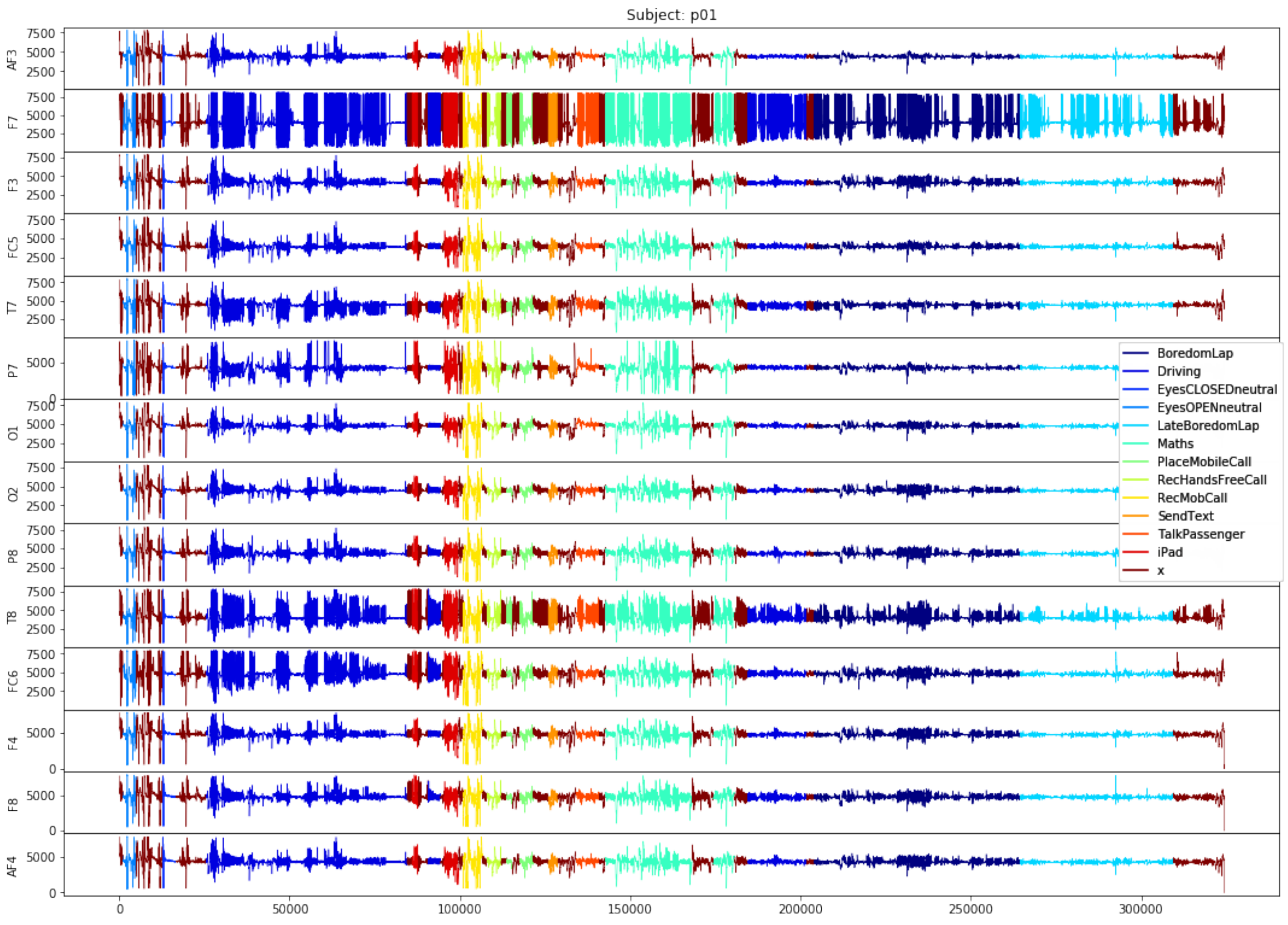}
\caption{Example of a weakly labelled EEG data}
\label{fig:eeg}
\end{figure}

Time-series classification \cw{(TSC)} is one of the most well-researched areas of modern computer science \cite{Bagnall2017-xw, dau2019ucr, Ismail_Fawaz2019-tc, lines2016hive, Dempster2020-ck, fawaz2019inceptiontime}. 
\cw{With the increase in time-series data, accurate real-time TSC has become more important than ever.}
Despite the huge progress in \cw{TSC} over the last decade \cite{Bagnall2017-xw,Dempster2020-ck,fawaz2019inceptiontime,lines2016hive,dau2019ucr}, most of these works make assumptions that are unrealistic in the real-world setting. 
This includes the assumption that the patterns (subsequences) that exist in the time-series are of fixed equal lengths and are always correctly labelled \cite{Hu2016-go}. 
However, in the real-world settings, these patterns can be of unequal length, the class label of the pattern may contain errors and may only correspond to a general section of the data \cite{inproceedings2}.
\cw{For example, detecting distracted driving using Electroencephalography (EEG) signal is a challenging task that violates the unrealistic assumptions required by most TSC algorithms.
Figure \ref{fig:eeg} shows an EEG signal}
obtained from a participant during driving using \cw{the} Emotiv Epoc+ EEG headset\footnote{\url{https://www.emotiv.com/}\label{emotivFootnote}}.

The 14\cw{-channel EEG signal can be interpreted as a time-series with 14 dimensions.
A time-series analysis algorithm can be applied to analyse these brain waves of the driver and detect the state of the driver to be either focused or distracted.} 
\cw{The different colors, as indicated in the legend, represent the different activities performed by the driver during driving.
These activities are weakly labelled because it is challenging to correctly label the driver's transition from driving to using an iPad and vice versa.
Besides, these subsequences may also contain noise or redundant patterns that are not related to the activity that the driver is performing at that point in time.
In fact, these redundant patterns may occur in different activity patterns, which makes it challenging to differentiate the different activities.}
\cw{Moreover, most real-world time-series dataset are highly imbalanced where they contain more of the ``normal'' label.
In this example, there are more driving labels than distracted labels, thus making the task of predicting a distracted activity extremely challenging.}

\cw{Quite simply, current state-of-the-art TSC algorithms are not capable of handling weakly labelled time-series data.}
It makes sense because these models were mainly designed
\cw{for univariate TSC \cite{Bagnall2017-xw,dau2019ucr}.}
While this strategy works well for noise-free time-series with proper
\cw{labelling}, it fails to extract discriminative features from weakly-labelled time-series as the signal-to-noise ratio is low.
Hence, most TSC algorithms \cite{Dempster2020-ck,Ismail_Fawaz2019-tc} cannot distinguish between noise and useful data, especially when the class labels are highly imbalanced.

This research
\cw{aims} to develop 
\cw{models} that can accurately classify multivariate weakly-labelled time-series. 
The approach explores the use of ``attention'' mechanisms \cite{bahdanau2014neural} to extract the contextual relevance of subsequences \cw{while minimising the effect of noise and redundancies.}
The intuition behind this is the chronological order of time-series, such that under realistic conditions, they follow certain chronological trends. 
For example, \cw{for the task of predicting distracted driving, the EEG data}
will typically show a driver who is distracted at one moment has a higher probability of being distracted in the next moment or show gradual changes, rather than a sudden spike in concentration.
Therefore, this research studies a set of deep learning models that can classify each subsequence of the time-series and understand sequence correlations using attention.

This paper is organised as follows. Section \ref{sec:background} discusses some background information and related work in state-of-the-art TSC algorithms and introduces existing attention strategies. Section \ref{sec:methodology} highlights the improvements made using a CNN-LSTM architecture to implement the idea of contextual relevance. It also describes the use of LSTM and attention mechanism. Finally, sections \ref{sec:experimentation} describe the experimental setup with the weakly labelled time-series evaluated in this research, followed by the discussion of improved results discussion obtained as a result of this study.

\section{Background and Related Work}
\label{sec:background}
The field of TSC has advanced significantly in the last decade, with each algorithm significantly more accurate than its predecessors \cite{Bagnall2017-xw,Dempster2020-ck,dau2019ucr,Ismail_Fawaz2019-tc,fawaz2019inceptiontime}.
This section outlines the strength and limitations of some of the existing state-of-the-art TSC algorithms \cw{used in this research} and some of the popular attention mechanisms \cite{Vaswani2017-xw,bahdanau2014neural,Luong2015-xr} available till date.

\subsection{State-of-the-art TSC Algorithms}
\label{subsec:sota tsc}
\cw{Recently,} Dempster et. al. \cite{Dempster2020-ck} proposed ROCKET, a \cw{TSC classifier that achieves state-of-the-art performances with a fraction of time of existing state-of-the-art methods.
ROCKET extract features from a time-series using a large number of random convolutional kernels. 
These kernels have random length, weights, bias, dilation, and padding, and when applied to a time-series produce a feature map.
Then the maximum value and the proportion of positive values are computed from each feature map, producing two real-valued numbers as features per kernel.
With the default 10,000 kernels, ROCKET produces 20,000 features.
These features are then used to train a ridge regression classifier.
ROCKET was found to be the most accurate TSC classifier compared with other state-of-the-art algorithms such as HIVE-COTE \cite{lines2016hive}, InceptionTime \cite{fawaz2019inceptiontime} and ResNet \cite{Ismail_Fawaz2019-tc} when bench-marked on the 85 TSC dataset \cite{dau2019ucr}.}
\cw{Although designed for univariate time-series,} it can easily be adapted for multivariate \cw{time-series} with similar performance. 
This research applies ROCKET on weakly-labelled multivariate time-series, however as shown later,
\cw{ROCKET does not perform well on weakly labelled time-series because it was designed using clean, well-labelled time-series dataset \cite{dau2019ucr}.}
In fact, none of the state-of-the-art TSC algorithms are designed to classify weakly labelled time-series.

Ismail et. al. \cite{Ismail_Fawaz2019-tc}
\cw{reviewed a few} 
deep-learning models \cw{for TSC}.
\cw{They found two models that achieve state-of-the-art performances,} Fully \cw{Convolutional} Network (FCN) and Residual Network (ResNet).
\cw{ResNet was ranked as the most accurate univariate TSC deep learning model bench-marked on 85 univariate TSC dataset \cite{dau2019ucr}. 
It is a convolutional neural network (CNN) that consists of three residual blocks with three convolutional layers in each block, followed by a global average pooling (GAP) layer and a softmax layer for classification.} With the help of residual connections, ResNet eliminates vanishing gradient effect\cite{Hochreiter1998-uf} in time-series where changes in data are very small (small learning gradient).

\cw{On the other hand, when bench-marked on 12 multivariate TSC dataset, FCN is the most accurate multivariate TSC deep learning model although it is not significantly different from ResNet \cite{Ismail_Fawaz2019-tc}. 
} 
FCN 
\cw{consists of} three \cw{1D} convolutional blocks \cw{with batch normalisation and ReLU activation function to learn features from the time series.} 
\cw{Then, a GAP} layer\cw{, applied to the last convolutional block summarises the features which are then passed to a softmax classifier for classification.} 
Perslev et. al. \cite{Perslev2019-db} 
\cw{proposed U-Time, a segmentation algorithm}
which uses a feed-forward network (FFN) with convolutional encoders and decoders to segment subsequences of long time-series. 
U-Time \cw{was designed for segmentation and} demonstrated high segmentation accuracy on a range of multivariate weakly labelled time-series such as real-life sleep-staging \cw{EEG} time-series. 
\cw{This research is interested in the classification of weakly labelled data. Hence U-Time is not being evaluated in this work and will be considered as future work.}

\cw{All of these state-of-the-art algorithms have shown superior performance on a set of benchmark dataset \cite{dau2019ucr}.
Most of these datasets are not realistic, i.e. they are properly labelled, equal in length and do not suffer from class imbalanced.
Hence these state-of-the-art algorithms are not capable of handling weakly labelled dataset with unequal length and highly imbalanced, such as the task of predicting distracted driving using EEG signal, as shown in our experiments in Section \ref{sec:experimentation}.}
Therefore, this research identifies the lack of \cw{suitable algorithms} 
to classify multivariate weakly-labelled time-series 
with high accuracy.

Among these existing state-of-the-art models, this research focuses on \cw{improving} FCN and ResNet because their neural network architecture allows them to be upgraded. 
\cw{More specifically, this paper explores various attention mechanisms to handle weakly labelled data.}

\subsection{Attention}

The attention concept was initially proposed in 2015 by Bahdanau et. al. \cite{bahdanau2014neural} to utilise the contextual information in sequence-to-sequence (Seq-2-Seq) problems such as natural language processing (NLP) \cite{bahdanau2014neural}. 
Seq-2-Seq models are neural networks that \cw{take in a sequence of data as input} and output \cw{another sequence, such as language translation.} 
These models typically \cw{consist of an encoder and a decoder \cite{chaudhari2019attentive}.}
\cw{The encoder tries to summarise the input sequence into a concise feature vector while the decoder maps the feature vector to an output sequence.
Using the example of language translation, the input can be an English sentence and the output can be a French sentence.}
When input sequences are long, generating the contextual information becomes a bottleneck. 
To prevent this problem, attention strategy was introduced to decouple the encoders and decoders, and allow long sentences to be translated accurately using alignment functions.

\subsubsection{Attention by Luong and Bahdanau et. al.}
Since the introduction of attention, several other designs were introduced by a number of subsequent papers, mostly focused on language translation.
In 2015, Luong et. al \cite{Luong2015-xr} proposed an attention model in a \cw{Long Short-Term Memory} (LSTM) encoder-decoder architecture.
Their model 
was based on the intuition to keep track of all the words in a sentence so that the relevance of each word to the current word can be used to make better translation decisions. 
Luongs' attention 
was an improvement to the previous strategy by  Bahdanau et. al. \cite{bahdanau2014neural}, because it uses multiple calculations to derive the context. However, the benefit of the model by Bahdanau et. al. \cite{bahdanau2014neural} was that it extracted a bi-directional sequence context.

\subsubsection{Self-Attention and Multi-Head Attention}
Later in 2017, a state-of-the-art attention model, known as Multi-Head Attention, was proposed by Vaswani et. al. \cite{Vaswani2017-xw} 
To learn the context in sequences, this model builds correlation functions using a neural network model known as Self-Attention and use a complex multi-layered encoder-decoder architecture, known as the Transformer \cite{Vaswani2017-xw}. This model can focus on multiple sections of sequences in parallel to extract context.

Since the introduction of these attention models, they have been adopted in  various research across different problem domains \cite{velivckovic2017graph}. In this research, some of the above-mentioned attention strategies \cw{are explored} in combination with the existing CNN TSC algorithms to demonstrate significant improvement in TSC.


\section{Methodology}
\label{sec:methodology}

\cw{This section outlines the approach taken in this study to improve existing TSC models on weakly labelled data.} The first sub-section discusses the intuition behind the use of context-relevance to improve weakly labelled TSC. 
Then the following sub-sections introduce the CNN-LSTM architecture implemented in this research and explain the implementation of existing attention strategies for complex context extraction in TSC. The final sub-section lists all the neural network models created in this study.

\subsection{Context relevance of subsequences}

Relevance information in time-series can be described as the additional context between subsequences. \cw{For example, if a subsequence currently being classified has the same class as the previous two subsequences, they are likely to have similar patterns that can help to improve TSC.}
When the \cw{models} incorrectly learn the features from subsequences due to noise and erroneous data, this additional context can help the models identify similar sections of a subsequence with previous subsequences to correctly identify its category.
This means \cw{that} even if the current subsequence is much longer or shorter in length, finding relevant sections from previously known subsequences can help the model to understand the correct category of the subsequence. Therefore, this idea eliminates the effect caused by subsequence length variations in TSC.

Similarly, if the data is imbalanced and the model does not have enough data to learn the features of the minority category, the contextual information can tip the classification decision for a subsequence towards the correct category as soon as it detects relevant sections with previous subsequences which also belong to that minority category.

Since time-series data \cw{are ordered in time,}
the level of noise or redundancy is expected to show continuous changes.
If a model can constantly keep track of the noise and redundancies by comparing the relevance of sections from previous subsequences, it can continuously learn and cancel out the effect of noise or redundancies. With this intuition, two aspects of a desired 
\cw{model} can be defined; it must be supervised to firstly allow the model to learn correct features corresponding to categories and secondly, it must be able to keep track of the contextual information for every subsequence with its nearest previous ones. 

Following the idea of context relevance, this study initially \cw{employs a CNN-LSTM structure to learn the relationship between subsequences.} 

\subsection{CNN-LSTM Architecture}

\cw{CNN-LSTMs have shown great success for similar tasks such as activity recognition, image description and video description \cite{donahue2015long}. 
The authors \cite{donahue2015long} investigated different combinations of CNNs and LSTMs to create models that are suitable for each of the tasks.}

\cw{Typically,} in a CNN-LSTM model, the CNNs are used to extract distinctive features from subsequences \cw{while LSTMs are used to learn the relationship between subsequences.} 
LSTM is a popular Recurrent Neural Network (RNN) and the benefit of using an RNN layer is its ability to ``remember from the past'' \cite{gers1999learning}. 
In particular, LSTM can be used to store context because it is influenced by both its given input weights and hidden-state vectors that represent context from previous inputs.

\begin{figure*}[ht]
\centering
\includegraphics[width=10cm]{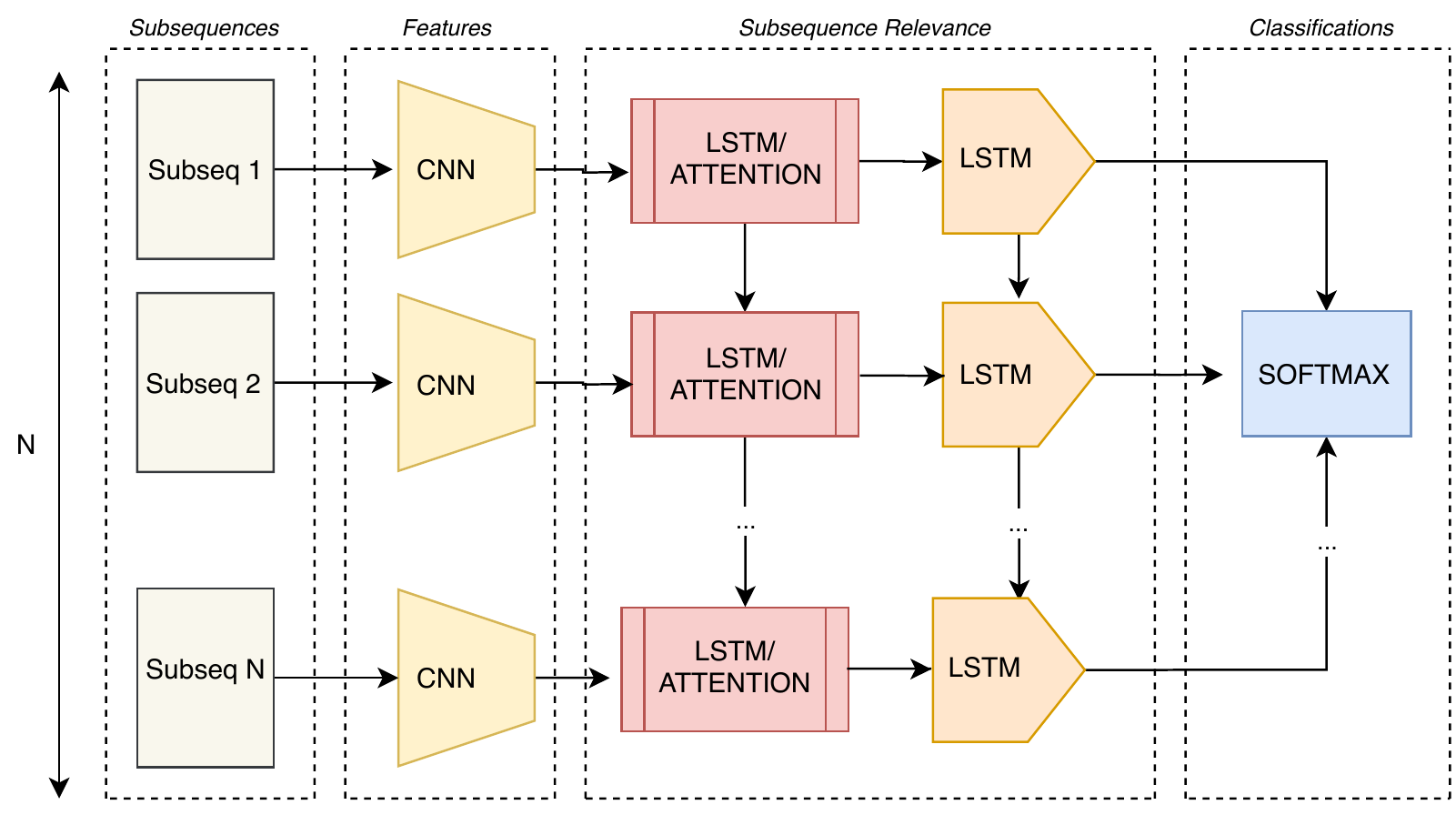}
\caption{\cw{A general} CNN-LSTM architecture, where $N$ is the number of subsequences.}
\label{fig:cnn-lstm}
\end{figure*}

\cw{In this study, state-of-the-art CNN TSC models are combined with two LSTM layers to form a CNN-LSTM architecture, illustrated in Figure \ref{fig:cnn-lstm}.}
Note that the CNN refers to either FCN or ResNet for TSC \cite{Ismail_Fawaz2019-tc} while LSTM or Attention (will be described in the next section) layers can be placed in between CNN and final LSTM for subsequence context extraction. This design is applicable to both univariate and multivariate time-series.

As illustrated in Figure \ref{fig:cnn-lstm}, $N$ number of subsequences can be provided to this CNN-LSTM. The first $N-1$ subsequences are used to extract context and when combined with the feature-maps produced from the CNN, it is expected that the final subsequence in the batch can be classified \cw{more accurately}.
Based on this architecture, this study creates two models: \textbf{FCN\_LSTM} and \textbf{ResNet\_LSTM}.

The intuition behind these models is that the state-of-the-art CNNs are able to extract discriminative features from subsequences which are then passed to the LSTMs. For example, when the CNN in Figure \ref{fig:cnn-lstm} is FCN \cite{Ismail_Fawaz2019-tc}, it contains 3 convolutional layers with 128, 256 and 128 filters respectively.
\cw{The LSTMs then extract context from the feature maps generated by the CNNs.}
The feature maps are then passed to two LSTM layers with 64 hidden units to extract context from the feature maps.
The output of the final LSTM layer is passed to a softmax layer for classification. 

Table \ref{tab:table1} shows the improvements made using CNN-LSTM. While this approach improves the classification performance when compared to existing CNNs, the performance can be further improved using more complex context relevance analysis. To explore this, attention mechanisms are added using an encoder-decoder architecture for context extraction.

\subsection{Attention for context extraction}

Generally, seq-2-seq models with attention use LSTM encoders and LSTM decoders \cite{li2019ea}. 
However, as shown in Figure \ref{fig:cnn-lstm}, this study implements Attention in a CNN-LSTM architecture between the CNN encoders and LSTM decoders,
because the CNNs \cw{have shown to achieve state of the art performances on various TSC tasks \cite{Ismail_Fawaz2019-tc}.} 

While LSTM in CNN-LSTM models can decode the contextual information from CNN feature-maps, attention mechanisms can be used to pay more ``attention'' to certain subsequences in the whole time series. 
Attention mechanisms use various techniques to align previous subsequences with the current subsequence to identify strong pattern matches. 

This research studies some state-of-the-art attention mechanisms.
\cw{Specifically, the following attention mechanisms are used to improve on the CNN-LSTM architecture mentioned in the previous section.} 

\begin{enumerate}[leftmargin=*]
  \setlength\itemsep{0.2em}
  \item Alignment Attention, introduced by Luong et. al. \cite{Luong2015-xr}
  \item Bi-directional LSTM with Attention, by Bahdanau et. al. \cite{bahdanau2014neural}
  \item Locally and Globally focused Self-Attention Model, designed by Vaswani et. al. \cite{Vaswani2017-xw}
  \item Multi-Head Attention Model, also by Vaswani et. al. \cite{Vaswani2017-xw}
\end{enumerate}

The method of how each attention mechanism extracts context relevance for TSC is explained across the next few subsections. It is important to note, while almost all of the attention algorithms discussed in this study are integrated with FCN and ResNet, some of them are implemented independently to evaluate their standalone performance.

\subsubsection{Luongs' Attention}






Since the original Luong et. al. \cite{Luong2015-xr} attention was designed for seq-2-seq models, this study implements the same architecture for time-series.

After receiving $N$ sets of feature-maps from the CNN encoders, Luongs' attention creates an alignment-score vector. These alignment scores represent how well each of the subsequences matches with the features extracted from the other subsequences. 

The alignment score vector is multiplied with the encoder output for the last subsequence which is a method of identifying how much of the first $N-1$ subsequences ``align'' with the last subsequence. In other terms, this dot product generates a context vector, that is the weighted sum of the feature-maps of the last subsequence which is relevant to the previous subsequences.


Theoretically,  the LSTM decoder can use these weights to understands how well the last sequence is relevant to the previous sequences. This is done by concatenating this context vector with the CNN feature-maps to update the contextual information to the feature-maps.

\subsubsection{Bi-directional Alignment Attention}

Similar to Luongs' strategy, the attention concept proposed by Bahdanau et. al. \cite{bahdanau2014neural} is implemented to extract the context of subsequences in forward and reverse direction. The intuition behind their algorithm is to understand the relevance of previous subsequences affecting the final one, and how the class of the previous subsequences could be affected based on the classification made on the final subsequence.



\begin{figure}[h]
\centering
\includegraphics[width=0.9\columnwidth]{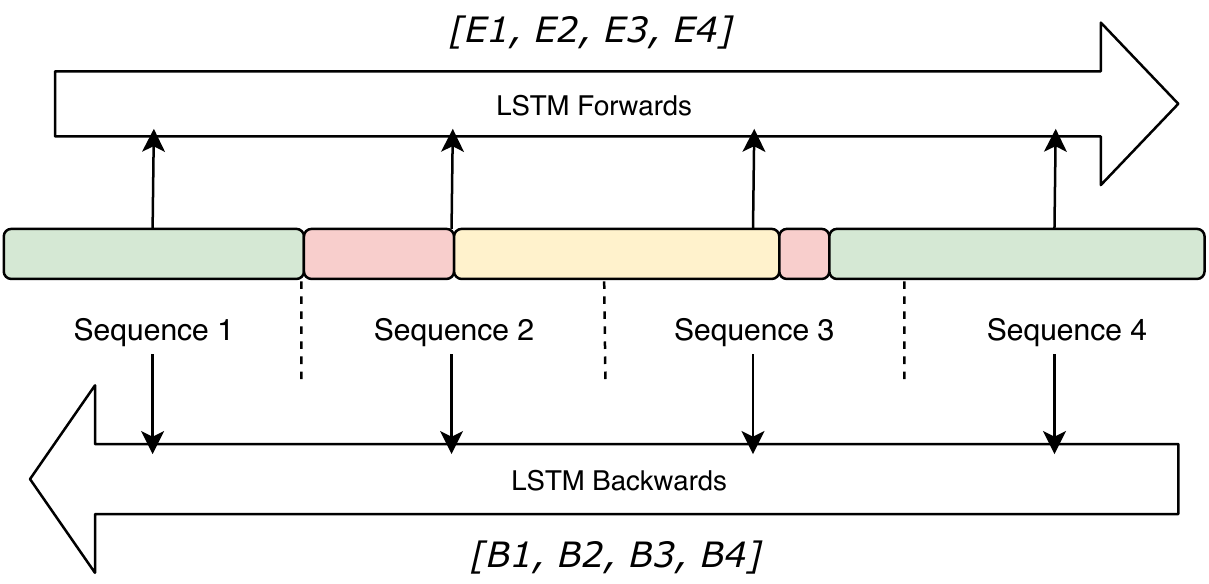}
\caption{Bi-directional LSTM with attention}
\label{fig:bidirectional}
\end{figure}

The model begins by receiving a batch of $N$ subsequence feature-maps. To allow analysis in both directions, this algorithm uses a bi-directional LSTM to learn the representation of the subsequences in order and backwards.

Figure \ref{fig:bidirectional} illustrates how each direction of the bi-directional LSTM create a set of hidden states, labelled $[E1, E2, E3, E4]$ for forward and $[B1, B2, B3, B4]$ for backwards when considering $N=4$ subsequences. Similar to Luongs' model \cite{Luong2015-xr}, these hidden states represent the alignment scores of subsequences individually for each direction and hence, provides much more context for TSC.

The algorithm then concatenates these hidden-states together, to allow each of them to collectively determine the overall alignment score of subsequences in both forward and backward directions. The weights of the vector after concatenation represent the context vector that is returned to the decoders.








\subsubsection{Self-Attention}

The Self-Attention model is made of complex ``Transformer'' encoders-decoders, which are series of Self-Attention layers and Feed-Forward Neural Networks \cite{Vaswani2017-xw}. The core Self-Attention (SelfA) library implemented in this study was adapted from a Keras implementation by CyberZHG\footnote{\label{CyberZHG}\url{https://github.com/CyberZHG}}.

The transformer generates and maintains Query(Q), Key(K) and Value(V) vectors \cite{Vaswani2017-xw}, which are responsible to keep track of the context of subsequences. After receiving feature maps from the CNN encoders, SelfA passes these through the Transformer encoders to calculate the alignment scores of subsequences.
Then these alignment scores are multiplied with its initially created value(V) vector with certain calculations to flush out irrelevant values from the context vector \cite{Vaswani2017-xw}.
In the case of weakly labelled time-series, this is expected to clean irrelevant context from the first three subsequences efficiently, to allow stronger matches to be found with the last subsequence.







The SelfA model also introduces the concept of ``soft'' vs. ``hard'' attention \cite{Vaswani2017-xw}.
In terms of TSC, soft attention creates a context vector of a window centred around the position of the current subsequence for maximum focus, and it stretches along the length of previous subsequences so the model gets an overall perspective of the context. On the contrary, hard attention aggressively focuses on specific areas of input subsequences and therefore, identifies the relevance of definite sections of the previous subsequences to the current. 

This study implements both global and local SelfA algorithms, where global SelfA uses soft attention to learn overall patterns across the previous subsequences, and local SelfA uses a balance of soft and hard attention to extract alignment score for subsequences.

This study also implements the two different alignment score functions proposed by the SelfA algorithm \cite{Vaswani2017-xw}. The additive function calculates the scores by passing the context vectors to a Feed-Forward Network (FFN), whereas the multiplicative method passes the vectors into a softmax layer where the scores are calculated as the following equation.

\begin{equation}Attention(Q, K, V)=softmax(\frac{QK^T}{\sqrt{d_k}})V\label{eq:1}\end{equation}

For context extraction of subsequences, the additive function is expected to outperform multiplicative function, especially when the encoder key dimension, d\textsubscript{k}, is large. This d\textsubscript{k} value depends on the complexity of time-series subsequences, that is, the higher the number of features of the subsequence, the higher the dimensions and vice versa. In the case of multiplicative approach, large d\textsubscript{k} means large magnitudes of alignment scores are produced and therefore, extreme gradient values are generated which would negatively impact context extraction. For weakly labelled time-series, the number of features extracted from subsequences is expected to be high particularly due to additional noise and redundancies. However, assuming that the complexity is not definitive, this research evaluates SelfA strategy with both multiplicative and additive functions.

The performance comparisons between global vs. local SelfA, and multiplicative vs. additive SelfA, are discussed later in Section \ref{sec:experimentation}.

\subsubsection{Multi-Head Attention}

Finally, one of the most advanced attention mechanisms, Multi-Head Attention (MHA), proposed by  \cite{Vaswani2017-xw}, is a strategy which adopts the idea of the previously discussed SelfA, but with more focused attention to various sections of previous subsequences in parallel.

This algorithm runs several self-attention models together to perform context extraction, which makes it computationally heavy. The MHA implemented in this study is primarily based on a Keras implementation by CyberZHG \footref{CyberZHG}, with improvements made from the original implementation, which is available as a Tensorflow library \footnote{\url{https://www.tensorflow.org/tutorials/text/transformer}}.


\begin{figure}[h]
\centering
\includegraphics[width=8cm]{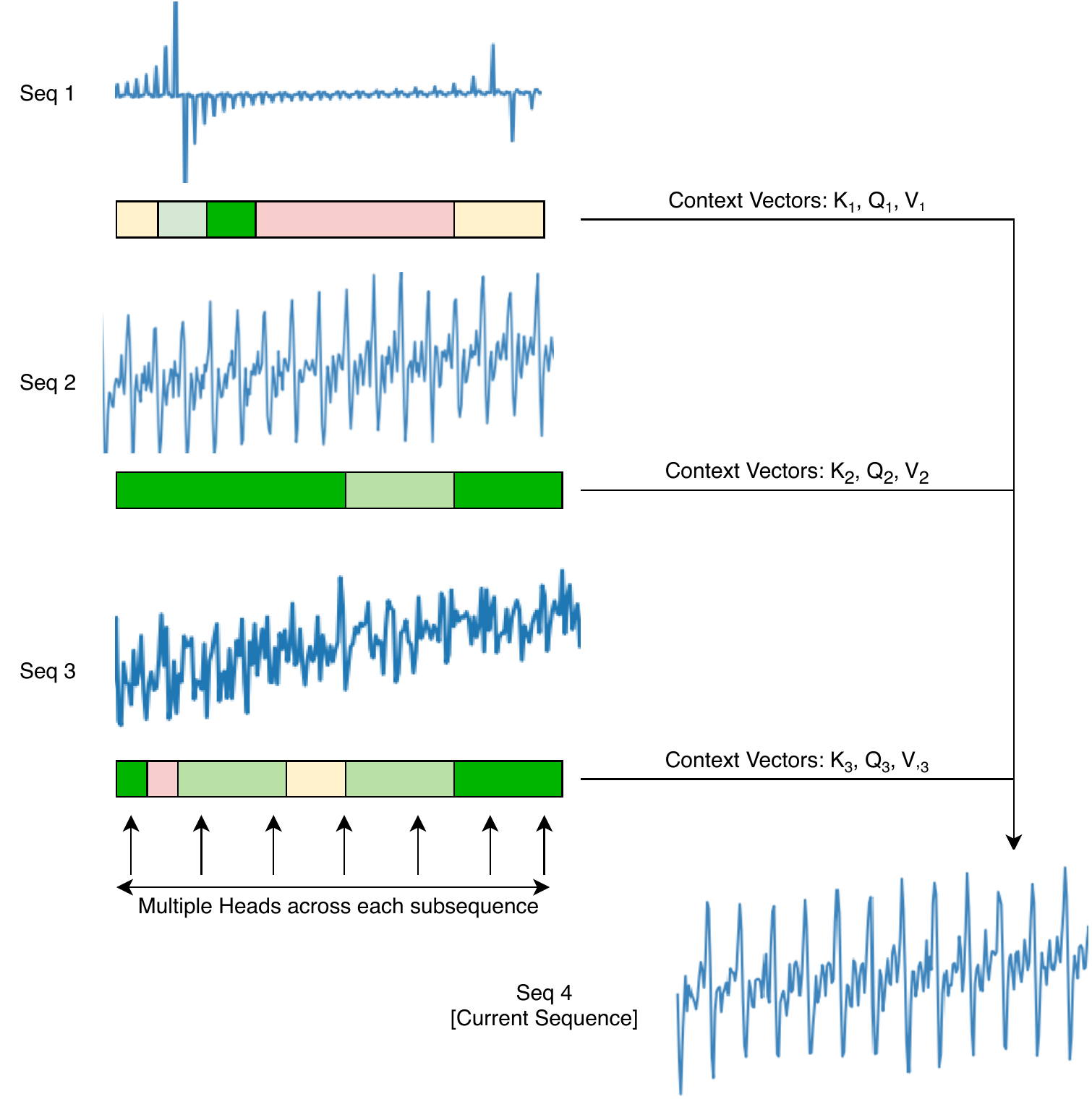}
\caption{An example of MHA focusing on different time-series subsequence positions}
\label{fig:MHA}
\end{figure}

From the CNN encoders, MHA receives a set of $N$ feature-maps per batch, again with the aim to classify the last one. This algorithm creates its initial vectors similar to SelfA but this is replicated ``h'' times, where h is the number of heads selected as a hyper-parameter.



With regards to TSC, MHA runs the parallel SelfA encoders to understand the relevance of multiple areas of input subsequence at the same time. Figure \ref{fig:MHA} illustrates how MHA determines that the last subsequence is relevant to the two immediate-previous subsequences, but not to the first one. The multiple heads in MHA allow the model to focus on individual positions within these two previous subsequences to compare the various sections with the subsequence, and at the same time flushing out the irrelevant sections using self-attention as discussed before.

While the multiple heads allow a parallel focus on previous subsequences, it is important that the model considers the order. For TSC, it is only useful to keep track of neighbouring subsequences if they are part of a running trend. Nevertheless, MHA takes care of the order using an additional vector known as the positional encoding. Theoretically, this keeps track of the distance between each extracted context locations, giving the model a sense of subsequence order.

MHA decoders are masked which means they will only consider previous subsequences. This is important in this study because it enables the model to suit real-time classifications where future subsequences are not known. While the previously discussed attention algorithms follow this restriction by default, MHA requires this additional masking due to multiple Self-Attention running in parallel and because MHA cannot centrally control the individual attention encoder/decoder positions.


\subsection{Multi-Step CNN-ATTENTION}

Since the CNN-LSTM encoder-decoder architecture is designed to receive $N$ subsequences per batch, this study labels these models as Multi-Step CNN-ATTENTION models, where $N>1$. The list of these models implemented and studied in this research are as follows.

\begin{itemize}
  \setlength\itemsep{0.2em}
  \item \textbf{FCN\_LATTN}: Luong’s Attention with FCN
  \item \textbf{ResNet\_LATTN}: Luong’s Attention with ResNet
  \item \textbf{FCN\_BATTN}: Bahdanau’s Attention with FCN
  \item \textbf{ResNet\_BATTN}: Bahdanau’s Attention with ResNet
  \item \textbf{SelfA}: Local Additive Self-Attention
  \item \textbf{FCN\_SelfA}: Local Additive Self-Attention with FCN
  \item \textbf{FCN\_SelfA\_Global}: Global Self-Attention with FCN
  \item \textbf{FCN\_SelfA\_Multiplicative}: Local Multiplicative Self-Attention with FCN
  \item \textbf{ResNet\_SelfA}: Local Self-Attention with ResNet
  \item \textbf{MHA}: Multi-Head Attention
  \item \textbf{FCN\_MHA}: Multi-Head Attention with FCN
  \item \textbf{ResNet\_MHA}: Multi-Head Attention with ResNet
\end{itemize}

\subsection{Single-Step CNN-ATTENTION}

The multi-step models extract the features and context from the past $N-1$ subsequences to predict the final one. However, to create a baseline of how the models perform when only one subsequence ($N=1$) is passed, a set of Single-Step CNN-ATTENTION models are created. 

The single-step models do not consider the relationship between subsequences, and will only consider the relevance of sections within a single sequence. For example, in the case of the EEG time-series, past three subsequences may show that the driver is continuously in a state of distraction and in the last subsequence, the first half show distraction trends, but the second half may start revealing increasing driving concentration. The single-step models are designed to disregard the relevance of the previous subsequence and evaluate the performance by only detecting context within single subsequences.

The single-step models comprise of FCN and ResNet combined with MHA. Only the MHA algorithm is selected because of its ability to focus on various positions of a subsequence using parallel attention ``heads''. Each of these heads are expected to apply the self-attention strategy to different positions within the single subsequence to extract the context. Similar to the multi-step models, the CNN layers learn the features of the subsequence while the MHA layer simply adds the context vector.

The single-step models implemented in this study are as follows.

\begin{itemize}[leftmargin=*]
  \setlength\itemsep{0.1em}
  \item \textbf{LSA}: Luong’s Single-step Attention
  \item \textbf{MHSA}: Multi-Head Single-step Attention
  \item \textbf{FCN\_MHSA}: Multi-Head Single-step Attention with FCN
  \item \textbf{ResNet\_MHSA}: Multi-Head Single-step Attention with ResNet
\end{itemize}

\section{Experimentation}
\label{sec:experimentation}

This section discusses the weakly labelled dataset used in this study. It also discusses the various performance metrics improved with the use of Attention for TSC.

\subsection{Experimental Setup}

All the models in this study are evaluated on two weakly labelled time-series datasets \texttt{Emotiv266} and \texttt{EmotivRaw}, supplied by the bioinformatics company Emotiv \footref{emotivFootnote}. The Emotiv datasets are collections of EEG brain signals data, captured using a light-weight and wireless Emotiv EPOC+ headset\footref{emotivFootnote}, from a 40-minute driving simulation by 18 participants performing 16 different tasks (defined in Figure \ref{fig:eeg} legend) while driving. 

\subsubsection{Details of the datasets by Emotiv}

The \texttt{EmotivRaw} dataset captures 14 individual EEG channels in parallel, at a frequency of 128Hz (illustrated in Figure \ref{fig:eeg}). As the participants perform various activities during driving, their brain activity alters and these are recorded in the EEG time-series. 
To remove ocular artifacts, a 4-40Hz band-pass filter and Fast Fourier Transform (FFT) \cite{Oberst2007-dh} was applied to every 2 seconds window with 0.25-second intervals. 
Each of the channels was further split into several frequency bands to extract features including average power, peak power and peak frequency from each of the frequency bands. 
Additional features were also created based on the aggregated data 
from various brain regions and occipital power within each specified band, and by accumulating EEP power. 
In total, 266 variables (channels) per EEG signal are created and combined to form \texttt{Emotiv266}.

\subsubsection{Data Preparation}

Among the 18 participants, 12 participants are randomly selected as the `training' set, another 2 as `validation' set and the remaining are used as `test' set to evaluate the models.

\begin{figure}[h]
\centering
\includegraphics[width=8cm]{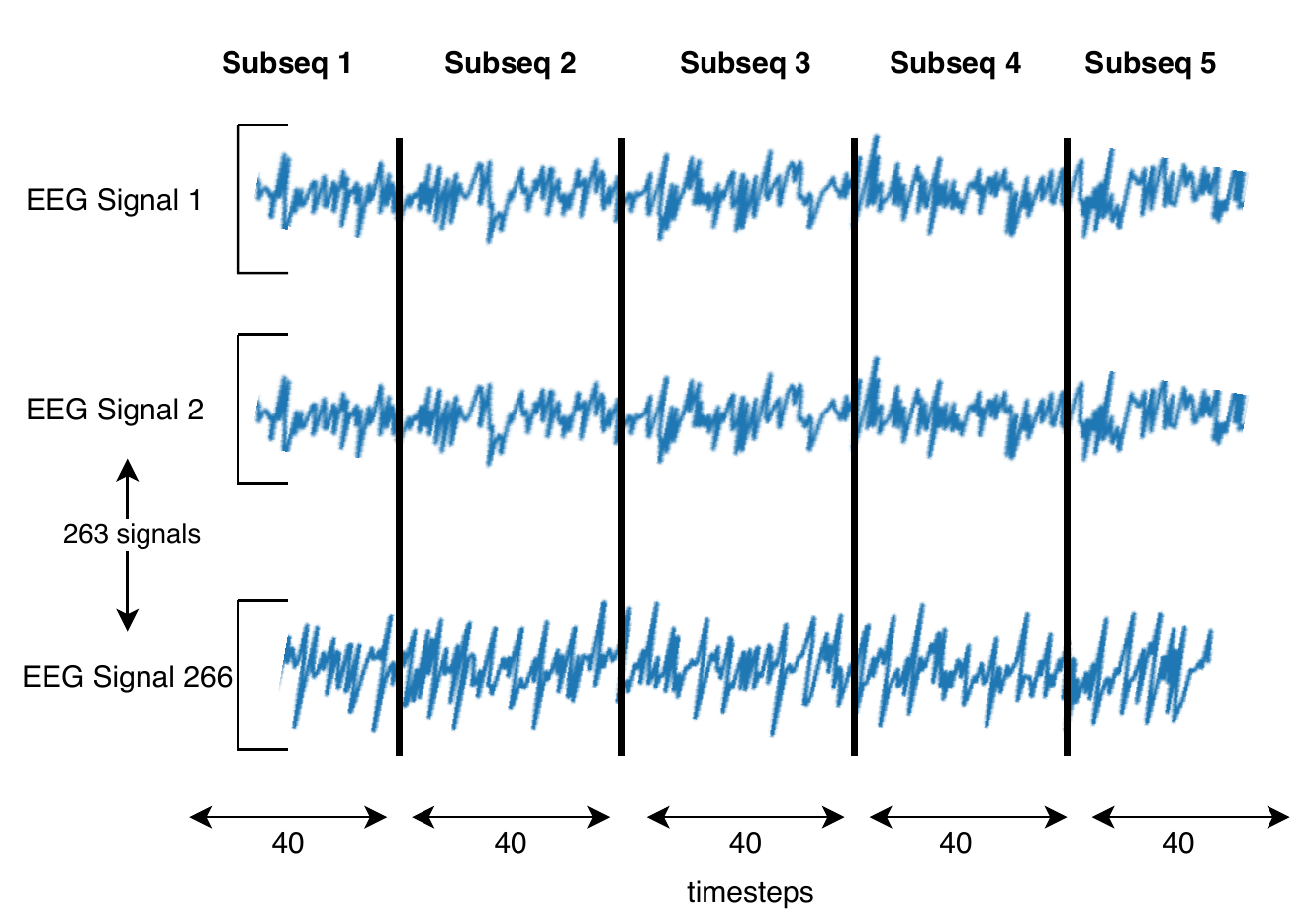}
\caption{Time-series segmentation of \texttt{Emotiv266}}
\label{fig:segmentation}
\end{figure}

\texttt{EmotivRaw} and \texttt{Emotiv266} are long multivariate time-series. To allow these time-series to be accepted by the deep-learning models, the long time-series is segmented into equal-length ``subsequences'', as shown in Figure \ref{fig:segmentation}.

The subsequences are extracted with a sliding window with 50\% overlap.
For \texttt{Emotiv266}, these segmented subsequences are of dimensions $266 \times 40$, which refers to 266 parallel time-series channels of 40 time-steps in length, corresponding to a 4.25 seconds window.
\cw{Similarly for \texttt{EmotivRaw} with 14 channels, sampled at 128Hz, the time series are segmented into subsequences of $14 \times 256$, representing 2 seconds window.}
To create a suitable simulation of a real-time stream, all the multi-step models are evaluated with 4 subsequences per batch, that is $N=4$. 
The four subsequences together represent a window of 11.75s (160 time-steps samples) \cw{for \texttt{Emotiv266} and 3.5s (1024 time-steps samples) for \texttt{EmotivRaw}.} 
The number of subsequences in training, validation and test set of \texttt{Emotiv266} are 7656, 995 and 2010 respectively, and in \texttt{EmotivRaw} are 18040, 2513 and 3947.


\subsubsection{Label Enhancements}

\cw{In this study, only considering binary classification are considered, where a class 0 represents distracted and class 1 represents focused driving respectively. Multi-class classification will be considered as future work where a more complex model is required.
Typically each of these subsequences are labelled with the majority label. This may introduce some unwanted signals in the subsequence which may lead to poor performance. Hence, a slightly different approach is considered in this study. If the whole subsequence does not contain multiple groups of labels, then the same label is assigned to the subsequence.}
However, if there are two different groups of labels across the subsequence, the subsequence is further split into two and each subsequence is filled with insignificant placeholder values to fill them to window sizes of 40. 
This ensure that each subsequence extracted has consistent labels. 
If the window has over two label variations (which is unlikely in Emotiv), the windows are discarded. 
This reduces any negative impact on TSC experimentation that can be caused by data preparation.

\subsubsection{Up-sampling}
\cw{Similar to most real-world datasets, the Emotiv datasets are also highly imbalanced.}
The majority label is ``focused'' driving, which is almost twice the number of ``distracted'' driving (a ratio of 3828 to 2032 samples).
Although the attention models are expected to mitigate the effect of imbalanced data, this study \cw{explores a simple method to balance out the classes in the data by random up-sampling the minority class in the training data before training. 
Note that other approaches such as SMOTE \cite{chawla2002smote} are not considered in this study as random up-sampling has found to work well in practice without the complexity of generating synthetic time series data.}
The effect of up-sampling data on the performance of the models of this study is analysed in detail later in Section \ref{sec:experimentation}.

The python library ``imbalanced-learn'' \footnote{\url{https://imbalanced-learn.readthedocs.io}} is used to repeatedly duplicate random records from the minority group (distracted category). 
This balances the number of records for each label and provides a fair learning opportunity for the models.

\subsection{Evaluating Performance}


\cw{All the models implemented in this study are trained with the categorical cross-entropy loss for 100 epochs and optimised with ADAM \cite{kingma2014adam}.
The learning rate is adjusted based on the loss gradient.}
This ensures the model can reach a minimum plateau in validation loss, and hence perform TSC with minimum loss. Moreover, the experiments are designed to minimise empirical risk \cite{taiwological} by selecting the best models with minimum loss among the epochs.
\cw{The code for this study has been made publicly available at \url{https://github.com/Surayez/EmotivDriverDistraction}.}



Table \ref{tab:table1} outlines the average test scores of each model on \texttt{Emotiv266} and \texttt{EmotivRaw} after five runs. 

\input{main_results_table}


\subsubsection{Accuracy}

Immediately noticeable from Table \ref{tab:table1}, most of the attention implementations perform with higher accuracies than the existing deep learning models. The addition of the LSTM layer, for basic context relevance, improved the accuracy scores of FCN and ResNet by almost 5\% on both datasets, which is slightly higher than the accuracy by achieved by ROCKET.

However, the most notable models are FCN\_SelfA, MHA and FCN\_MHA, which show significant improvements in classification accuracy. On average, FCN\_SelfA, classifies approximately 13\% more accurately compared to the existing FCN model and over 7\% than FCN\_LSTM on \texttt{Emotiv266}. Again on \texttt{EmotivRaw} FCN\_SelfA shows 7\% and 4\% enhancement over FCN and FCN\_LSTM. Following this, FCN\_MHA and MHA perform with over 10\% higher accuracies than FCN, and 4\% higher accuracies when compared to FCN\_LSTM on \texttt{Emotiv266}, and almost 7\% and 3\% on \texttt{EmotivRaw}.
These show dramatic improvements in weakly labelled TSC, in comparison to the existing models.

In fact, the accuracy scores of SelfA and MHA by themselves are surprisingly high. It indicates that the position based context extraction of SelfA and MHA can solely perform TSC more accurately on weakly labelled data than the existing models. However, the high loss metrics of these two models, discussed later, indicate that this is not always the case.

The simpler LATTN and BATTN models also demonstrate improved TSC accuracy. In comparison to existing models, FCN\_LATTN and FCN\_BATTN are almost 5\% more accurate on both dataset, but when compared with FCN\_LSTM the performance is almost identical. This indicates that the additional attention in these models are not improving the performance significantly. This is largely due to underfitting, as discussed in more detail in Section \ref{sec:underfitting}.

\vspace{\baselineskip}


\subsubsection{AUC}

To further establish the improvement using attention algorithms, the AUC scores are evaluated as part of this analysis. AUC scores are important for this binary classification evaluation as it eliminates the effect of bias in the target categories when comparing the average performance of classification algorithms \cite{Huang2005-bn}. 

The CNN-ATTENTION models clearly demonstrate significant improvements in AUC metrics. FCN\_MHA performs with an almost 5\% higher AUC score compared to the FCN model, which has the highest AUC score among the existing TSC models. Similarly, FCN\_SelfA shows approximately 3\% improvement compared to FCN on both datasets. Even when LSTM extracts the basic context, it improves the AUC by almost 2\% on \texttt{Emotiv266}. This demonstrates that the higher accuracy scores of FCN are, in fact, improving the overall performance of TSC.

On the other hand, the ResNet models show much lower AUC scores in comparison to FCN models. This indicates that the accuracy improvements when using attention with ResNet may not always be improving the overall performance of the models. The primary reason for this is that ResNet is overfitting the training data, and this issue is discussed in detail later in section \ref{sec:overfitting}.

\subsubsection{F1 Score}
The F1 scores of each model are important in this study because the initial Emotiv data is highly imbalanced \cite{inproceedings}. Table \ref{tab:table1} shows the maximum average F1 score among the existing models is at 54\% achieved by ROCKET. Although FCN\_SelfA and FCN\_MHA improved the F1 score by almost 4\% when compared to FCN on \texttt{Emotiv266} and 2\% on \texttt{EmotivRaw}, this is not a drastic improvement considering the F1 score by ROCKET is around the same.

However, the performance of CNN-ATTENTION in terms of F1 score is reflected when the models are evaluated with imbalanced data. Table \ref{tab:table2} compares the F1 and accuracy scores by the existing FCN and ResNet, and the CNN-ATTENTION model with the highest accuracy from Table \ref{tab:table1}, which is FCN\_SelfA, before and after the training set of \texttt{Emotiv266} is up-sampled.

\input{upsample-results}

Table \ref{tab:table2} shows that the accuracy and F1 driving scores of FCN and ResNet are quite high when the data is imbalanced. However, the F1 distracted score is significantly low which means that the models can accurately classify most of the driving (dominant) subsequences, but only a few of the distracted (minor) category. The overall accuracy is high due to the higher true positives where in reality, most of the distracted category classifications are incorrect.

When the data is up-sampled (balanced), the number of accurate distracted classifications improve, hence the distracted F1 scores rise sharply. Now that FCN and ResNet are equally false-classifying for both categories, the overall accuracy falls. Similar behaviour is observed on \texttt{EmotivRaw}.

Nevertheless, the F1 scores of FCN\_SelfA show that the effect of up-sampling has a minor effect on both the accuracy and F1 scores for driving. Moreover, the difference between the distracted F1 scores before and after up-sampling is much smaller than in the case of FCN and ResNet. This proves that the attention strategy, as expected, is assisting the FCN model to make better decisions of the minority category when data is imbalanced. 
Using the context information from previous subsequences, it is able to eliminate the effect of imbalanced data. 
Therefore, it can be concluded that the improved accuracy of FCN\_SelfA would be preserved even when the input time-series is imbalanced, as in a real-world setting.

\subsubsection{FCN and ResNet}

The performance evaluations of this study indicate that ResNet has inconsistent improvements when combined with LSTM or attention algorithms. Table \ref{tab:table1} shows that ResNet performs well when combined with SelfA and BATTN, but show lower improvements with MHA and LATTN. The overall accuracy metrics suggest that FCN is performing significantly better with additional context information than ResNet.

However, the reason for this behaviour can be pointed to the more complex architecture of ResNet. The feature-maps returned from 3 convolution blocks of FCN seems to be better interpreted by the attention, than with that from 9 convolution blocks of ResNet. Because of this, ResNet overfits the training set slightly more than FCN and hence, lower accuracies are reflected in Table \ref{tab:table1}.

\subsubsection{Single-Step Baseline Models}

When the single-step models are evaluated on both the Emotiv dataset, they did not demonstrate exceptional performance metrics. The results from Table \ref{tab:table1} indicate that the single-step models slightly improve TSC when compared to FCN and ResNet, but not as much as the multi-step models.

This confirms the intuition of this research that the multi-step models are indeed using subsequence relevance to increase accuracy. It also suggests that understanding the context of various positions within the single subsequence is not enough to minimise the effect of weakly labelled data in Emotiv time-series. The attention models examining a larger window for context (multi-step models) are performing much better. 

However, these performance metrics do not disqualify the single-step models from weakly labelled TSC. Intuitively, the models are expected to prove very useful when subsequences of time-series are of higher lengths and show a more repetitive pattern. For the case of \texttt{Emotiv266} or \texttt{EmotivRaw}, each subsequence block is not exceptionally long to take advantage of such context analysis and therefore, the single-step models show insignificant improvements compared to the existing models.

\subsubsection{Validation Loss}

\begin{figure}[h]
\centering
\includegraphics[width=0.8\columnwidth]{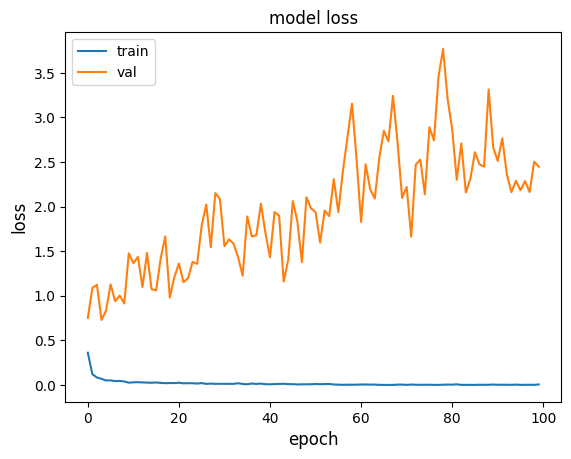}
\caption{Training and validation loss of FCN\_SelfA}
\label{fig:epoch_SelfA_FCN}
\vspace{-5pt}
\end{figure}

Although there is an increase in overall accuracy, the loss graphs generated by the FCN\_SelfA (shown in Figure \ref{fig:epoch_SelfA_FCN}) and FCN\_MHA models indicate overfitting of the training set. This is identified by the increasing and fluctuating validation loss while the training loss decrease exponentially.

Due to overfitting, when the models are given an unseen data (validation set), the false predictions are heavily penalised by significantly increasing loss and correct predictions have a minor effect on decreasing the loss \cite{Salman2019-at}. This explains the behaviour of Self-Attention and MHA classifying with high accuracy but also with higher loss.

\begin{figure}[h]
\centering
\includegraphics[width=0.8\columnwidth]{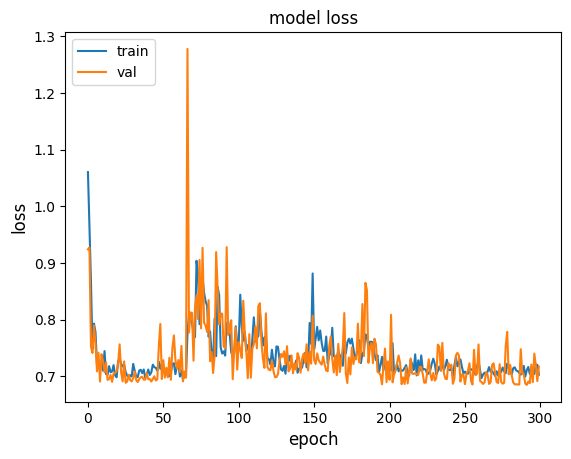}
\caption{Training and validation loss of FCN\_BATTN}
\label{fig:epoch_fcn_BATTN}
\end{figure}

On the contrary, LATTN and BATTN (shown in Figure \ref{fig:epoch_fcn_BATTN}) models have a better loss pattern despite their minor improvements in accuracy. Each of its validation loss shows an exponential decrease, similar to its training loss. Once again, this is a typical indication that the models are underfitting the training data and is discussed in section \ref{sec:underfitting}.

\subsubsection{Effect of Overfitting}
\label{sec:overfitting}

\begin{figure}[h]
\centering
\includegraphics[width=8cm]{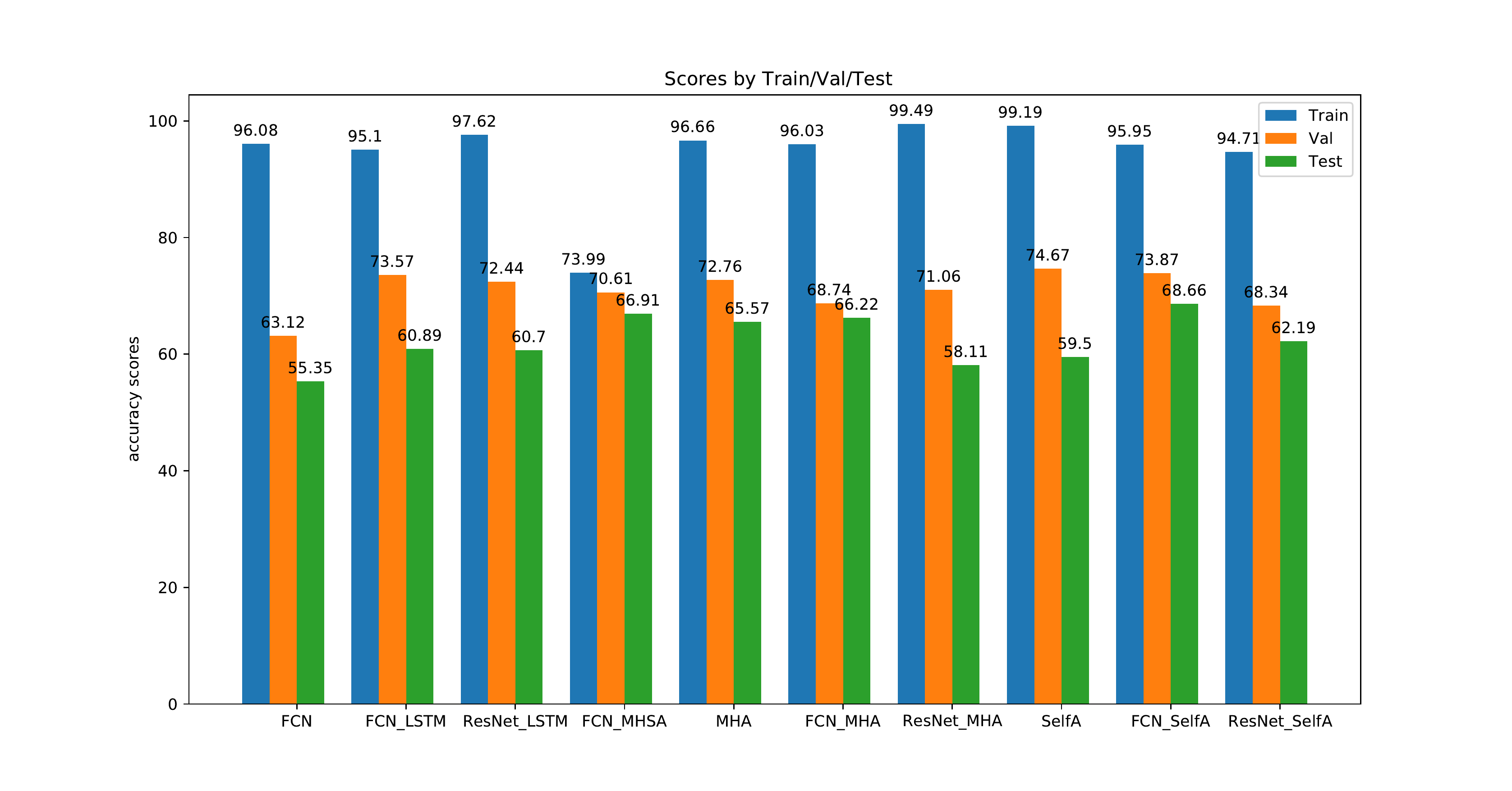}
\caption{Accuracy of TSC models with 100 epochs}
\vspace{-5pt}
\label{fig:tsc-accuracy}
\end{figure}

Figure \ref{fig:tsc-accuracy} shows the training, validation and test accuracies in a sample evaluation of the models on the \texttt{Emotiv266} time-series. 
It is clear that almost all the models have a much higher training accuracy than test accuracy, which identifies that the models are largely overfitting the training set \cite{jabbar2015methods}.

The reason for this behaviour is due to the complex architectures of FCN and ResNet. 
When these models are trained on a weakly labelled dataset before attention is applied, they learn the noise and redundancies of the training data too accurately. This clouds their judgement of the noise as distinctive features of subsequences. As the models are then evaluated on test data, they fail to identify the new patterns of noise and redundancies in test set, leading to low test set accuracies.

As expected, the attention layers significantly improve the accuracy of the models by cutting down this effect. The improved results indicate that attention can successfully support TSC by extracting context of subsequences. However, this effect is not completely eliminated by attention.

\subsubsection{Effect of Underfitting}
\label{sec:underfitting}

Attention algorithms by Luong et. al. \cite{Luong2015-xr} and Bahdanau et. al. \cite{bahdanau2014neural} are considering subsequence relevance information because it shows minor improvements, but the simplicity of the alignment calculation algorithm fails to retrieve enough context in comparison to SelfA or MHA. The more complex internal encoders and decoders of SelfA and MHA provide them with powerful computational support to 'attend’ more sections within the previous subsequences, and therefore better alignments to be found. This explains the underfitting behaviour of LATTN and BATTN which have low accuracies despite having a low validation loss metric.

Nevertheless, LATTN and BATTN are re-evaluated in this study on longer epochs (300 epochs) with the expectation that further training may eliminate the effect of underfitting. 
Unfortunately, the performance did not improve even with the longer training period.

\subsubsection{Global vs. Local SelfA}

SelfA is one of the best performing models of this study and further analysis are conducted to explore various aspects of SelfA for TSC. Firstly, this research evaluates the performance of global and localised SelfA algorithms on \texttt{Emotiv266}. 
As discussed earlier, local SelfA is expected to be more accurate for \texttt{Emotiv266} because it considers various positions within previous subsequences, rather than a global ``soft'' focus. The results are consistent with this expectation because on average, local SelfA performed 4\% more accurately than global SelfA. Similar performance is observed on \texttt{EmotivRaw}.

\subsubsection{Additive vs. Multiplicative SelfA}

Furthermore, the performance of additive and multiplicative SelfA are evaluated on \texttt{Emotiv266}. When local FCN\_SelfA is evaluated with the additive alignment function, it performed 5\% more accurately than when evaluated with multiplicative alignment function.

The reason for this performance difference is the previously discussed FFN layer, which is present in the additive algorithm but is replaced with a softmax calculation in Equation \eqref{eq:1} in the multiplicative algorithm. Due to the noise and data redundancies, the subsequences in Emotiv are of high complexities. The extra FFN layer helps the additive model interpret the encoder keys with larger dimensions of the Emotiv subsequences, and causes the multiplicative calculation to produce extreme gradient values which negatively impact context extraction.

\subsubsection{Effect of normalisation}

\input{normalisation_results}

Table \ref{tab:table3} shows the average outcomes when the models are evaluated on time-series with:
\begin{itemize}[leftmargin=*]
  \setlength\itemsep{0.05em}
  \item No normalisation
  \item Only normalising the subsequences
  \item Only normalising the time series
  \item Both normalisation as above
\end{itemize}

While it is a standard practice to normalise time-series for TSC using deep-learning \cite{lei2020time}, the performance of the models without the two types of normalisation applied in this study is evaluated to further reinforce the TSC improvements using attention. The time-series are normalised when loading the dataset and subsequences are normalised during segmentation. The outcomes in Table \ref{tab:table3} highlights the improved performance using attention by both Self-Attention and Multi-Head Attention models on \texttt{Emotiv266}. Exact patterns are observed for \texttt{EmotivRaw}

\section{Conclusion}
From the experiments on the real-world Emotiv dataset, it can be concluded that the attention mechanism improves the performance of existing state-of-the-art TSC algorithms on weakly-labelled time-series. To eliminate the effect of noise, redundancies and data imbalances, as in a real-world setting, the extraction and interpretation of subsequence context information help supervised CNNs to improve their performance in TSC. In particular, the performance of Self-Attention with FCN (with additive function in localised context) and Multi-Head Attention with FCN displays the most dramatic improvements, and can therefore be concluded as the top models of this study.

The set of models developed as part of this research can complement any further studies for real-time TSC. The CNN-LSTM architecture can be used to further analyse the performance of these models on datasets from different domains.

\vspace{\baselineskip}



%
\bibliographystyle{abbrv}
\bibliography{main}  

\end{document}

%% file: packages.tex
\usepackage{enumitem}
\usepackage{graphicx}
\usepackage{booktabs, siunitx}
\usepackage[svgnames,table]{xcolor}
\usepackage[tableposition=above]{caption}
\usepackage{pifont}
\usepackage{url}
\usepackage{todonotes}
\usepackage[hidelinks]{hyperref}
\usepackage{pdfpages}
\usepackage{fixltx2e}
\usepackage{footmisc}

%% file: main_results_table.tex
\begin{table*}[ht]
\centering
\rowcolors{7}{White}{LightBlue!30}
\renewcommand{\arraystretch}{1.0}
\begin{tabular}{lllllll}
        \toprule
        \multicolumn{1}{c}{} &
        \multicolumn{3}{c}{Emotiv266} & 
        \multicolumn{3}{c}{EmotivRaw}\\
        \cmidrule(lr){2-4}
        \cmidrule(lr){5-7}
        & {Accuracy} & {AUC} & {F1 Distracted} & {Accuracy} & {AUC} & {F1 Distracted} \\
        Models & {(\si{\percent})} & {(\si{\percent})} & {(\si{\percent})} & {(\si{\percent})} & {(\si{\percent})} & {(\si{\percent})} \\
        \midrule
Existing Models             &           &                  &                  &           &                  &                 \\
        \midrule
ROCKET                       & 59.49     & 41.09            & 54.1            & 59.21     & 41.38            & 54.73           \\
FCN                          & 55.06     & 58.64            & 50.96           & 57.3      & 50.55            & 50.6            \\
ResNet                       & 53.24     & 54.59            & 45.75           & 56.99     & 40.02            & 53.32           \\
        \midrule
\rowcolor{white} LSTM Models            &           &                  &                  &           &                  &                 \\
        \midrule
FCN\_LSTM                    & 60.23     & 60.31            & 48.57           & 60.94     & 49.81            & 53.2            \\
ResNet\_LSTM                 & 61.51     & 53.43            & 48.72           & 60.69     & 53.12            & 46.12           \\
        \midrule
Single-Step CNN-ATTENTION Models &           &                  &                 &           &                  &                 \\
        \midrule

LSA                          & 55.34     & 40.21            & 48.67           & 56.82     & 41.1             & 47.39           \\
MHSA                         & 61.88     & 54.56            & 59.62           & 60.47     & 43.03            & 46.97           \\
FCN\_MHSA                    & 63.91     & 58.51            & 46.96           & 62.63     & 44.78            & 47.55           \\
ResNet\_MHSA                 & 62.83     & 53.91            & 40.01           & 60.11     & 40.12            & 44.29
  \\
  	\midrule
 \rowcolor{white} Multi-Step CNN-ATTENTION Models  &           &                  &                 &           &                  &                 \\
        \midrule

SelfA                        & 60.95     & 58.2             & 42.1            & 61.65     & 42.09            & 48.57           \\
MHA                          & 64.32     & 60.95            & 50.01           & 62.19     & 48.51            & 39.47           \\
\textbf{FCN\_SelfA}                   & \textbf{68.32}     & \textbf{61.94}            & \textbf{54.71}           & \textbf{64.12}     & \textbf{52.36}            & \textbf{54.55}           \\
\textbf{FCN\_SelfA\_Global}           & \textbf{65.81}     & 60.43            & 52.29           & 62.07     & 50.88            & 52.71            \\
\textbf{FCN\_MHA}                     & \textbf{66.16}     & \textbf{63.51}            & \textbf{54.21}           & \textbf{64.25}     & 47.11            & 52.39           \\
FCN\_LATTN                   & 60.4      & 56.56            & 48.66           & 59.29     & 40.44            & 49.23           \\
FCN\_BATTN                   & 60.76     & 59.01            & 50.63           & 59.92     & 44.37            & 46.91           \\
ResNet\_MHA                  & 61.11     & 54.84            & 41.39           & 60.83     & 52.09 
    & 48.36\\
ResNet\_SelfA                & 62.21     & 60.4             & 54.4            & 59.87     & 42.85            & 54.1            \\
ResNet\_LATTN                & 54.99     & 48.7             & 50.85           & 58.91     & 39.01            & 52.06           \\
ResNet\_BATTN                & 63.19     & 50.4             & 52.83           & 60.56     & 42.09            & 53.8            \\
\bottomrule
\end{tabular}
\caption{Performance of each model on both Emotiv dataset, averaged over 5 runs. The ones in bold indicate the best performing models.}
\vspace{-15pt}
\label{tab:table1}
\end{table*}

%% file: upsample-results.tex
\begin{table}[h]
\centering
\rowcolors{3}{White}{LightBlue!30}
\renewcommand{\arraystretch}{1.0}
\begin{tabular}{lllll}
        \toprule
        & {Up-} & {F1 Distr.} & {F1 Driving} & {Accuracy} \\
        Models & {sampled} & {(\si{\percent})} & {(\si{\percent})} & {(\si{\percent})} \\
        \midrule
FCN        & No             & 22.47                 & 79.88              & 66.51           \\
           & Yes             & 50.96                 & 60.25              & 55.06           \\
ResNet     & No             & 20.58                 & 80.62              & 65.27           \\
           & Yes             & 45.49                 & 62.73              & 53.24           \\
FCN\_SelfA & No             & 35.21                 & 77.21              & \textbf{65.55}           \\
           & Yes             & 54.71                 & 74.89              & \textbf{68.32}           \\
\bottomrule
\end{tabular}
\caption{Performance variations based on up-sampling}
\vspace{-5pt}
\label{tab:table2}
\end{table}

%% file: normalisation_results.tex
\begin{table}[h]
\centering
\rowcolors{6}{White}{LightBlue!30}
\renewcommand{\arraystretch}{1.0}
\begin{tabular}{lllll}
        \toprule
        \multicolumn{1}{c}{} &
        \multicolumn{4}{c}{Accuracy with normalisation:} \\
        \cmidrule(lr){2-5}
        & {None} & {Subsequence} & {Series} & {Both} \\
        Models & {($\%$)} & {($\%$)} & {($\%$)} & {($\%$)} \\
        \midrule
FCN          & 69.29    & 63.12 & 54.32 & 55.06 \\
ResNet       & 65.54    & 64.42 & 52.16 & 53.24 \\
FCN\_LSTM    & 70.12    & 67.78 & 70.64 & 60.23 \\
ResNet\_LSTM & 67.36    & 66.41 & 64.51 & 61.51 \\
\textbf{FCN\_SelfA}   & \textbf{71.96}    & \textbf{68.09} & \textbf{71.29} & \textbf{68.32} \\
FCN\_MHA     & 71.13    & 67.26 & 70.35 & 66.16 \\
\bottomrule
\end{tabular}
\caption{Normalisation Analysis}
\vspace{-8pt}
\label{tab:table3}
\end{table}